\documentclass{article}

\usepackage[preprint,nonatbib]{neurips_2025}

\usepackage[utf8]{inputenc} 
\usepackage[T1]{fontenc}    
\usepackage{hyperref}       
\usepackage{url}            
\usepackage{booktabs}       
\usepackage{amsfonts}       
\usepackage{nicefrac}       
\usepackage{microtype}      
\usepackage{xcolor}         
\usepackage{amsmath}
\usepackage{graphicx}
\usepackage{tcolorbox}
\usepackage{xcolor}
\usepackage{tabularx}
\tcbuselibrary{skins}
\usepackage{multirow}
\definecolor{darkblue}{rgb}{0, 0, 0.5}
\hypersetup{colorlinks=true, citecolor=darkblue, linkcolor=darkblue, urlcolor=darkblue}
\title{Walk Before You Run!\\ Concise LLM Reasoning via Reinforcement Learning}

\author{%
  Mingyang Song, Mao Zheng \\
  Tencent Hunyuan\\
  \texttt{nickmysong@tencent.com} \\
}

\begin{document}
\maketitle
\begin{abstract}
As test-time scaling becomes a pivotal research frontier in Large Language Models (LLMs) development, contemporary and advanced post-training methodologies increasingly focus on extending the generation length of long Chain-of-Thought (CoT) responses to enhance reasoning capabilities toward DeepSeek R1-like performance. However, recent studies reveal a persistent overthinking phenomenon in state-of-the-art reasoning models, manifesting as excessive redundancy or repetitive thinking patterns in long CoT responses. To address this issue, in this paper, we propose a simple yet effective two-stage reinforcement learning framework for achieving concise reasoning in LLMs, named \textbf{ConciseR}. Specifically, the first stage, using more training steps, aims to incentivize the model's reasoning capabilities via \textbf{G}roup \textbf{R}elative \textbf{P}olicy \textbf{O}ptimization with \textit{clip-higher} and \textit{dynamic sampling} components (\textbf{GRPO++}), and the second stage, using fewer training steps, explicitly enforces conciseness and improves efficiency via \textbf{L}ength-aware \textbf{G}roup \textbf{R}elative \textbf{P}olicy \textbf{O}ptimization (\textbf{L-GRPO}). \textbf{{Significantly, \textbf{ConciseR} only optimizes response length once all rollouts of a sample are correct, following the "walk before you run" principle.}} Extensive experimental results demonstrate that our \textbf{ConciseR} model, which generates more concise CoT reasoning responses, outperforms recent state-of-the-art reasoning models with zero RL paradigm across AIME 2024, MATH-500, AMC 2023, Minerva, and Olympiad benchmarks. The code, training dataset, and model checkpoints will be publicly released\footnote{\url{https://github.com/nick7nlp/ConciseR}}.
\begin{figure}[h]
	\begin{center}
		\includegraphics[scale=0.335]{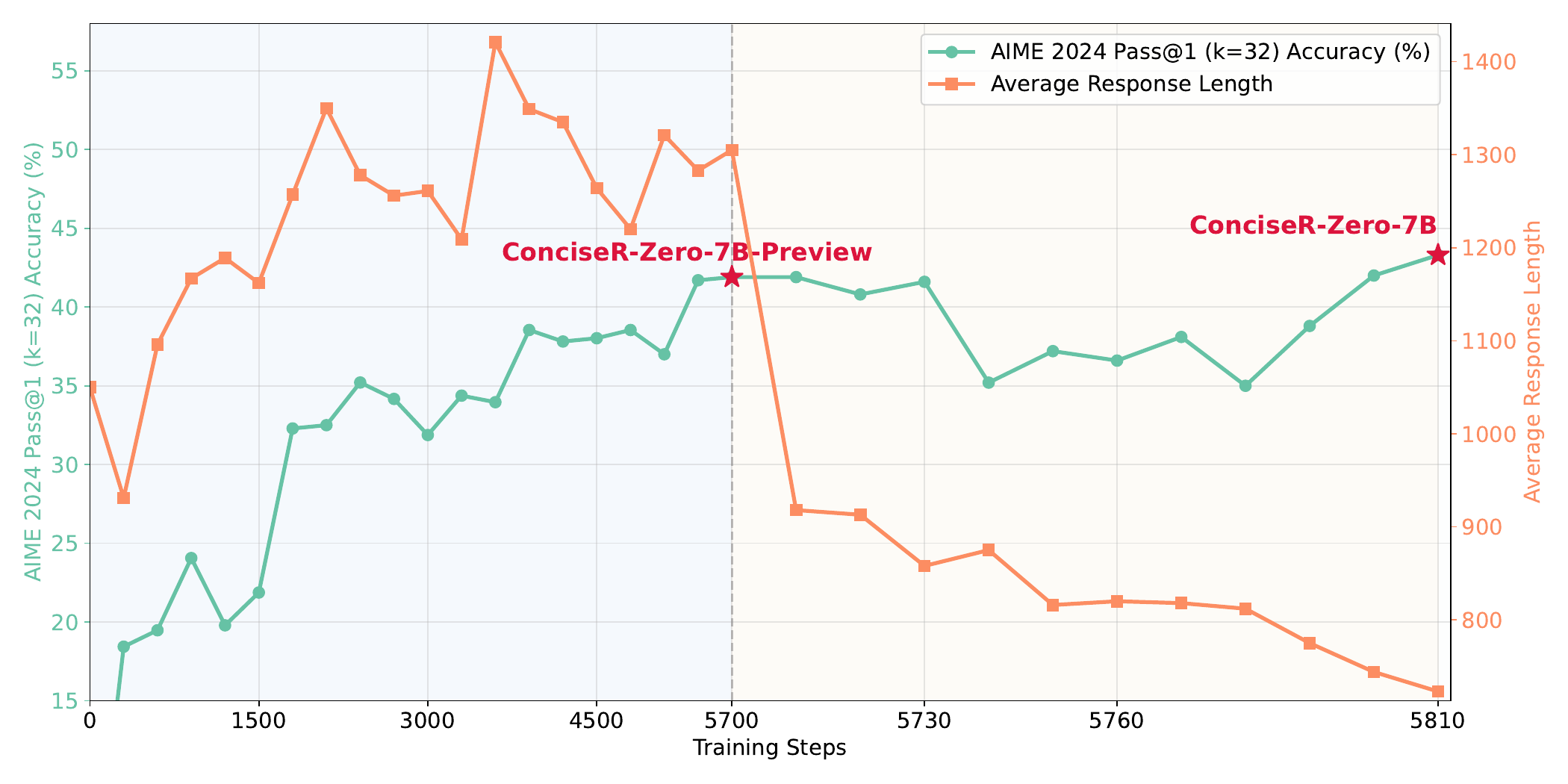}
	\end{center}
	\caption{A detailed evaluation of accuracy and response length throughout the training steps.}\label{curves}
\end{figure}
\end{abstract}

\section{Introduction}\label{introduction}
Test-time scaling~\cite{tts,muennighoff2025s1} has demonstrated a robust correlation between extending the generation length of Chain-of-Thought (CoT)~\cite{cot} and improving the reasoning capabilities of Large Language Models (LLMs). The advent of large reasoning models, such as GPT-o1~\cite{o1} and DeepSeek-R1~\cite{r1}, represents a significant breakthrough in natural language processing, especially in tackling complex and intricate reasoning tasks. An interesting phenomenon observed during reinforcement learning post-training via Group Relative Policy Optimization (GRPO)~\cite{grpo} is the emergence of an "aha moment"~\cite{r1}, which refers to a pivotal inflection point at which the model spontaneously initiates self-correction behaviors. These emergent behaviors develop autonomously through the model's exploration of the solution space rather than through explicit programming. Prior research~\cite{yeo2025demystifyinglongchainofthoughtreasoning,deepscaler} has found a distinctive pattern following this moment: the response length of the model tends to increase significantly, accompanied by improvements in overall performance. Despite the lack of a clear understanding of why this occurs, this phenomenon has led many researchers to advocate for longer responses, leveraging additional computational resources in the hope of further enhancing accuracy.

However, generating excessively long CoT reasoning responses substantially increases computational overhead in both the model training and deployment phases. Furthermore, recent studies~\cite{kimik1.5, cuadron2025dangeroverthinkingexaminingreasoningaction} have discovered an intrinsic overthinking phenomenon in reasoning models, where these models persistently produce verbose rationalizations. This tendency manifests as the inclusion of irrelevant contextual information and unnecessary reflective behaviors. Such information and behaviors not only inefficiently consume computational resources but also compromise reasoning accuracy by causing models to deviate from valid logical pathways to incorrect conclusions.

To address these issues, recent studies~\cite{deepscaler,fastcurl,dapo,drgrpo,fatemi2025concise} are researching efficient reasoning methodologies based on the GRPO algorithm for training the model to produce more concise CoT responses, and have discovered a trade-off between the CoT response length and model reasoning capabilities in most cases, i.e., the shorter the length, the worse the performance.
It is understandable that achieving efficient reasoning, improving ability via a more concise CoT, is inherently more challenging. This is in contrast to boosting performance by merely increasing the response length, as the former requires significantly higher model capabilities. Therefore, we highlight the critical importance of the timing for optimizing response length when training using GRPO-based algorithms. Adhering to the "walk before you run" principle, we consider that during training, response length optimization is only enabled when all rollouts for a training sample are correct.

Motivated by this, we propose ConciseR, which is a simple yet effective two-stage reinforcement learning framework for concise reasoning. Specifically, the first stage aims to enhance the model's reasoning capabilities through group relative policy optimization with clip-higher and dynamic sampling techniques. Meanwhile, we introduce the entropy bonus in the first stage, which is used to encourage exploration in policy gradient to incentivize the reasoning capabilities of the model further. The second stage enforces conciseness and improves efficiency via length-aware group relative policy optimization (incorporating max response length to calculate the length reward). Experiments demonstrate that ConciseR is compatible with RL models that incentivize reasoning, achieving reductions in response length while improving accuracy across benchmarks such as AIME 2024, AMC 2023, MATH-500, Minerva, and Olympiad datasets. As shown in Figure~\ref{curves}, ConciseR gradually activates the reasoning ability of the model in the first stage, then rapidly compresses the CoT response length in the second stage to achieve concise reasoning. Notably, the reduction in response length has immediate implications for computational efficiency, resource utilization, and response time, making the approach practically appealing and cost-effective.

\section{Preliminary}\label{preliminary}
\subsection{Proximal Policy Optimization (PPO)}
Proximal Policy Optimization (PPO)~\cite{ppo} is one of the policy gradient methods that introduces a clipped surrogate objective for policy optimization. By using clipping to constrain policy updates within a proximal region of the previous policy, PPO stabilizes training and improves sample efficiency. Specifically, PPO updates the policy by maximizing the following objective:
\begin{equation}
	\begin{aligned}
	\mathcal{J}_{\mathrm{PPO}}&(\theta) = \mathbb{E}_{{q} \sim \mathcal{D}, {o}_{\leq t} \sim \pi_{\theta_{\mathrm{old}}}(\cdot|{q})}\\&\min \left[ \frac{\pi_{\theta}(o_t \mid q, o_{<t})}{\pi_{\theta_{\mathrm{old}}}(o_t \mid q, o_{<t})} \hat{A}_t, \; \mathrm{clip}\left(\frac{\pi_{\theta}(o_t \mid q, o_{<t})}{\pi_{\theta_{\mathrm{old}}}(o_t \mid q, o_{<t})}, 1-\varepsilon, 1+\varepsilon\right) \hat{A}_t \right],
	\end{aligned}
\end{equation}
where $\pi_{\theta_{\mathrm{old}}}$ is the policy before the update, $\varepsilon$ is the clipping hyper-parameter, $q$ indicates the question from the data distribution $ \mathcal{D}$, and $\hat{A}_t$ is an  estimator of the advantage function of the $t$-th token. Here, a standard and traditional way to estimate $\hat{A}_t$ is to compute the Generalized Advantage Estimation (GAE)~\cite{ppo} with a learned value model. However, in the context of LLM RL scaling, learning the value model is computationally expensive, so methods that estimate $\hat{A}_t$ without a learned value model are practically preferred. 
\subsection{Group Relative Policy Optimization (GRPO)}
Group Relative Policy Optimization (GRPO)~\cite{grpo} introduces a policy gradient framework that eliminates the reliance on explicit value function by utilizing comparative advantage estimation within a group of responses. This method samples multiple candidate outputs for each input question and computing advantages based on the relative rewards among these candidates within their respective groups. Specifically, GRPO first samples a group of responses $\{o_1, o_2, \ldots, o_\text{G}\}$ per question and computes their returns $\mathbf{r} = \{r_1, r_2,  \ldots, r_\text{G}\}$, then calculates the advantage of the $i$-th response $o_i$ as,
\begin{equation}
	\hat{A}_i = \frac{r_i - \text{mean}(\{r_1, r_2, \cdots, r_{\text{G}}\})}{\text{std}(\{r_1, r_2, \cdots, r_{\text{G}}\})}.
\end{equation}
Similar to PPO, GRPO uses a clipped objective with KL penalty and optimizes the policy model $\pi_\theta$ by maximizing the following objective:
\begin{equation}
	\begin{aligned}
		\mathcal{J}_{\text{GRPO}}(\theta) = &\mathbb{E}_{{q} \sim \mathcal{D}, \{{o}_i\}_{i=1}^{\text{G}} \sim \pi_{\theta_{\text{old}}}(\cdot|q)} \\
		&\frac{1}{\text{G}}\sum_{i=1}^{\text{G}}\left\{\min\left[\tau_i(\theta)\hat{A}_i, \text{clip}\left(\tau_i(\theta), 1-\varepsilon, 1+\varepsilon\right)\hat{A}_i\right] - \beta\mathbb{D}_{\text{KL}}[\pi_\theta||\pi_{\text{ref}}]\right\},
	\end{aligned}
\end{equation}
where 
\begin{equation}
	\tau_i = \frac{\pi_\theta(o_i|q)}{\pi_{\theta_{\text{old}}}(o_i|q)}, \quad \mathbb{D}_{\text{KL}} (\pi_\theta \| \pi_{\text{ref}}) = \frac{\pi_{\text{ref}}(o_i|q)}{\pi_\theta(o_i|q)} - \log \frac{\pi_{\text{ref}}(o_i|q)}{\pi_\theta(o_i|q)} - 1.
\end{equation}
Here, $\pi_{\text{ref}}$ represents the reference model and the term $\mathbb{D}_{\text{KL}} (\pi_\theta \| \pi_{\text{ref}})$ indicates a KL penalty term to limit how much the trained model $\pi_\theta$ can deviate from the reference model $\pi_{\text{ref}}$.
\subsection{Decouple Clip and Dynamic Sampling Policy Optimization (DAPO)}
An in-depth analysis~\cite{dapo} reveals that the naive GRPO baseline suffers from several significant issues, such as entropy collapse, reward noise, and training instability. To address this issue, DAPO introduces four key techniques to make RL shine in the long-CoT RL scenario, including 
\begin{itemize}
	\item \textit{Clip-Higher}, which enhances the diversity of the model and avoids entropy collapse.
	\item \textit{Dynamic Sampling}, which improves training efficiency and stability.
	\item \textit{Token-Level Policy Gradient Loss}, which plays a crucial role in reinforcement learning scenarios involving Long-CoT reasoning responses.
	\item \textit{Overlong Reward Shaping}, a length-aware penalty mechanism designed to shape the reward for truncated samples to reduce reward noise and stabilize training.
\end{itemize}
Similar to GRPO, DAPO estimates the advantage in a group-relative manner and optimizes the policy model via the following objective,
\begin{equation}
	\begin{aligned}
		\mathcal{J}_{\text{DAPO}}(\theta) = &\ \mathbb{E}_{{q} \sim \mathcal{D}, \{{o}_i\}_{i=1}^{\text{G}} \sim \pi_{\theta_{\text{old}}}(\cdot|q)} 
		\frac{1}{\text{G}}\sum_{i=1}^{\text{G}}\left\{\min\left[\tau_i(\theta)\hat{A}_i, \text{clip}\left(\tau_i(\theta), 1-\varepsilon_l, 1+\varepsilon_h\right)\hat{A}_i\right]\right\}\\
		& \text{s.t.} \quad0 <| \{o_i | \text{is\_equivalent}(o_i, a)\} | < \text{G},
	\end{aligned}
\end{equation}
where $\varepsilon_l$ and $\varepsilon_h$ indicate the lower and higher clipping range.
\section{Methodology}\label{methodology}
Our primary goal is to let LLM generate a more concise CoT response without sacrificing the model's performance. To this end, we propose a novel two-stage reinforcement learning training paradigm guided by the principle of "\textit{Aim for 100\% accuracy first; speed comes with mastery.}" Specifically, the first stage aims to incentivize the reasoning capabilities of the base model via group relative policy optimization with clip-higher and dynamic sampling, thus ensuring accuracy and robust model reasoning. Subsequently, the second stage explicitly enforces conciseness and improved efficiency via length-aware group relative policy optimization, aligning with our objective of achieving mastery through precision, where conciseness naturally follows from reliable and accurate reasoning.
\subsection{Group Relative Policy Optimization with Clip-Higher and Dynamic Sampling (GRPO++)}
In the first stage, the model is trained to incentivize the reasoning capabilities, which aims to enhance the model’s problem-solving capacity, with an expected increase in response length as GRPO mostly encounters negative rewards and encourages the trained model toward longer responses. 
To this end, in this paper, we adopt GRPO with two key components of DAPO, clip higher and dynamic sampling, and further introduce an entropy bonus to encourage greater exploration capability in the model, named GRPO++. Similar to the original approach, GRPO++ estimates the advantage in a group-relative manner and optimizes the policy model using the following objective:
\begin{equation}
	\begin{aligned}
		\mathcal{J}_{\text{GRPO++}}&(\theta) = \mathbb{E}_{{q} \sim \mathcal{D}, \{{o}_i\}_{i=1}^{\text{G}} \sim \pi_{\theta_{\text{old}}}(\cdot|q)} \\
		&\frac{1}{\text{G}}\sum_{i=1}^{\text{G}}\left\{\min\left[\tau_i(\theta)\hat{A}_i, \text{clip}\left(\tau_i(\theta), 1-\varepsilon_l, 1+\varepsilon_h\right)\hat{A}_i\right] + \alpha\mathbb{H}(\pi_\theta)\right\},\\
		& \text{s.t.} \quad0 <| \{o_i | \text{is\_equivalent}(o_i, a)\} | < \text{G},
	\end{aligned}
\end{equation}
where $\alpha\mathbb{H}(\pi_\theta)$ denotes the entropy bonus.

\begin{table*}[h]
	\caption{Analyze the character-level output length and the frequency of reflections of correct and incorrect responses for \textsc{DeepSeek-R1-Distill-Qwen}-1.5B and -7B on AIME 2024.}\label{length}
	\begin{center}
		\resizebox{\textwidth}{!}{
			\begin{tabular}{lcccccc}
				\toprule
				\multirow{2}{*}{Model} &\multicolumn{3}{c}{\# Average Output Length} &\multicolumn{3}{l}{\# Average Frequency of "Wait" and "wait"}\\
				&\multicolumn{1}{c}{\textsc{Total}} &\multicolumn{1}{c}{\textsc{Correct}} &\multicolumn{1}{c}{\textsc{Incorrect}}&\multicolumn{1}{c}{\textsc{Total}} &\multicolumn{1}{c}{\textsc{Correct}} &\multicolumn{1}{c}{\textsc{Incorrect}}\\
				\midrule
				\textsc{DeepSeek-R1-Distill-Qwen}-1.5B &43176&21859&52629&109&49&138\\
				\textsc{DeepSeek-R1-Distill-Qwen}-7B &34193&19315&52376&75&35&124\\
				\bottomrule
		\end{tabular}}
	\end{center}
\end{table*}

\begin{figure}[h]
	\begin{center}
		\includegraphics[scale=0.265]{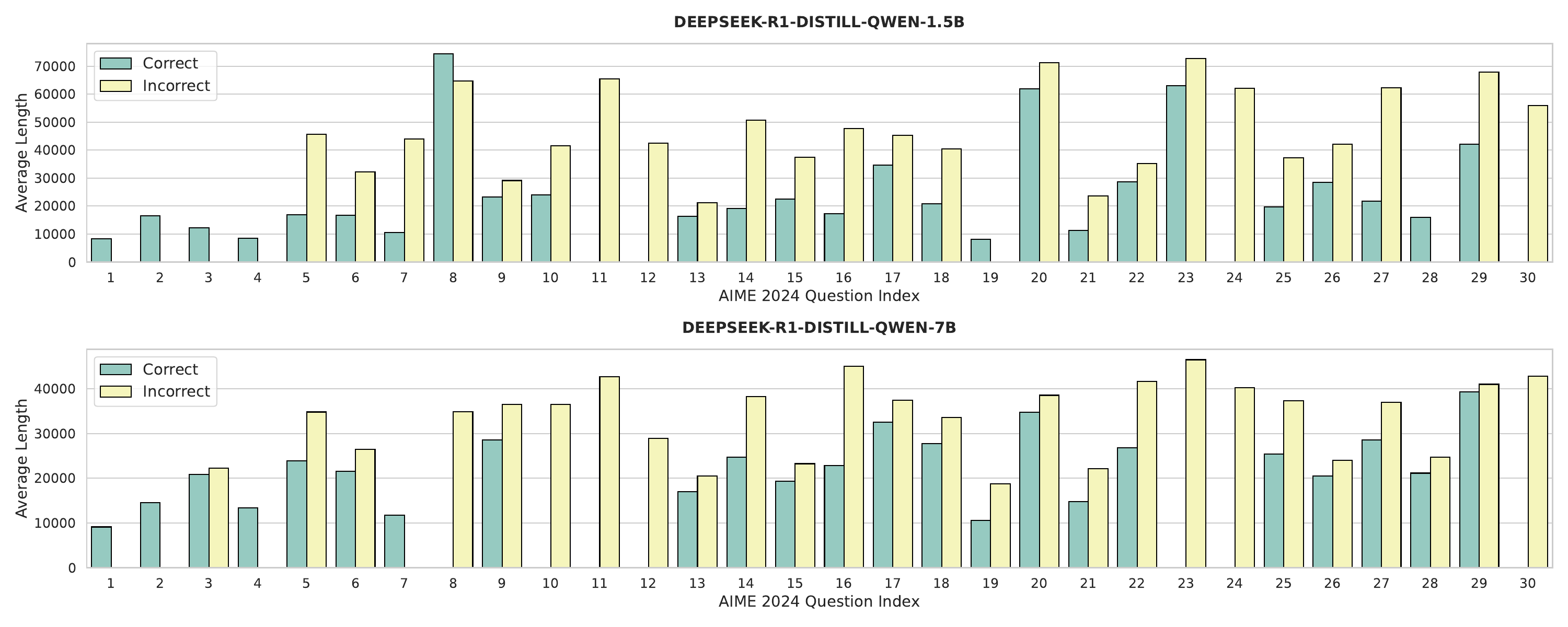}
	\end{center}
	\caption{Distribution of average length for correct and incorrect answers under the same question.}\label{vertical}
\end{figure}

\subsection{Length-Aware Group Relative Policy Optimization (L-GRPO)}
\begin{figure}[t!]
	\begin{center}
		\includegraphics[scale=0.23]{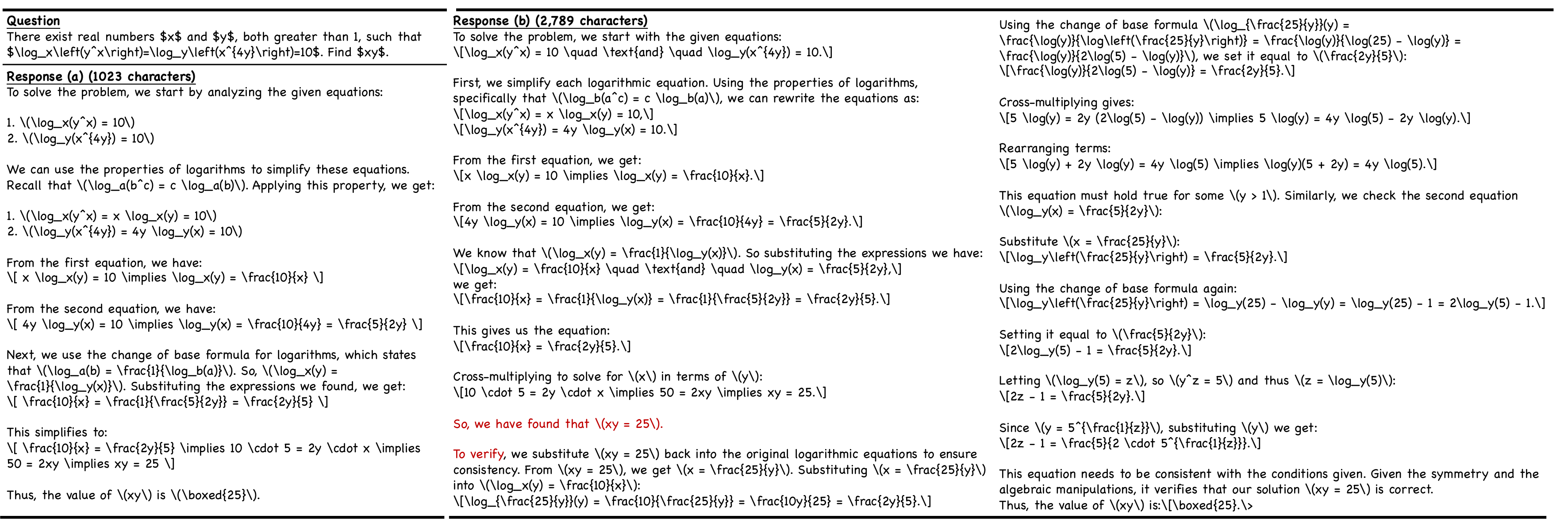}
	\end{center}
	\caption{Comparison of responses of ConciseR-Zero-7B-Preview to the same question.}\label{compare}
\end{figure}

Recent studies~\cite{deepscaler,fastcurl} have found that the reasoning response length is not strongly correlated with the correctness of the answer; that is, a long CoT reasoning response does not necessarily represent a correct result, and a short CoT reasoning response does not necessarily represent an incorrect one. On the contrary, the correct CoT reasoning responses are usually shorter in length, while incorrect reasoning responses tend to be longer. Therefore, we first analyze the response lengths of \textsc{DeepSeek-R1-Distill-Qwen}-1.5B and -7B, as shown in Table~\ref{length} and Figure~\ref{vertical}.
Table~\ref{length} shows that incorrect responses are obviously longer and include more detailed reasoning processes.
Simultaneously, to differentiate whether the incorrect results all stemmed from complex problems, thereby causing the CoT for incorrect responses to be longer in the statistical results, we analyze the model's responses at the sample level, as shown in Figure~\ref{vertical}. Interestingly, in most cases, for the same question, correct responses are still shorter. Furthermore, we conduct a deeper analysis of the correct responses to the same question and find that correct answers may exhibit excessive reflection, which leads to long CoT, resulting in longer CoT responses, as shown in Figure~\ref{compare}. 

Based on the above analysis, we reshape the reward function in GRPO. When the model's rollout results for a question are all correct, we further optimize the model's reasoning length for that question by using the remaining maximum response length as a reward (under the specified context length, the more remaining context length, the higher the reward), as calculated below.
\begin{equation}
	\tilde{A}_i = \frac{\hat{r}_i - \text{mean}(\{\hat{r}_1, \hat{r}_2, \cdots, \hat{r}_{\text{G}}\})}{\text{std}(\{\hat{r}_1, \hat{r}_2, \cdots, \hat{r}_{\text{G}}\})}, \quad\hat{r}_i = r_i + \lambda\hat{\mathcal{L}}_i, \quad \hat{\mathcal{L}}_i = 
	\begin{cases}
		1 - \frac{\mathcal{L}_i}{\mathcal{L}_{\text{Max}}}, & \text{if}\ \sum_{i=1}^{\text{G}}r_i = \text{G} \\
		0, & \text{if}\ \sum_{i=1}^{\text{G}}r_i \neq \text{G}
	\end{cases}
\end{equation}
where $\mathcal{L}_i$ is the length of the $i$-th response and $\mathcal{L}_{\text{Max}}$ indicates the max response length. Then, we optimize the policy model using the following objective:
\begin{equation}
	\begin{aligned}
		\mathcal{J}_{\text{L-GRPO}}&(\theta) = \mathbb{E}_{{q} \sim \mathcal{D}, \{{o}_i\}_{i=1}^{\text{G}} \sim \pi_{\theta_{\text{old}}}(\cdot|q)} \\
		&\frac{1}{\text{G}}\sum_{i=1}^{\text{G}}\left\{\min\left[\tau_i(\theta)\hat{A}_i, \text{clip}\left(\tau_i(\theta), 1-\varepsilon_l, 1+\varepsilon_h\right)\hat{A}_i\right] - \beta\mathbb{D}_{\text{KL}}[\pi_\theta||\pi_{\text{ref}}] + \alpha\mathbb{H}(\pi_\theta)\right\},\\
		& \text{s.t.} \quad0 <| \{o_i | \text{is\_equivalent}(o_i, a)\} |.
	\end{aligned}
\end{equation}
\subsection{Rule-based Reward Model}
Using a trained reward model typically introduces the issue of reward hacking [24–29]. To mitigate this issue, we directly adopt the final accuracy from a verifiable task as the outcome reward, calculated according to the following rule:
\begin{equation}
	r_i(o_i, a) = 
	\begin{cases}
		1, & \text{if}\ \text{is\_equivalent}(o_i, a) \\
		0, & \text{if}\ \text{not is\_equivalent}(o_i, a) \\
	\end{cases}
\end{equation}
where $a$ indicates the ground-truth answer and $o_i$ contains the predicted answer. Additionally, it is important to note that the trained model must adhere strictly to the training prompt by generating the chain-of-thought within the <think></think> tags and subsequently presenting the final answer within the <answer></answer> tags with the boxed tag.

\begin{table}[t]
	\centering
	\caption{Training Template. \textbf{\{{question\}}} will be replaced with the specific question during training.}
	\begin{tabular}{l}
		\toprule
		A conversation between User and Assistant. The user asks a question, and the Assistant solves it. \\
		The assistant first thinks about the reasoning process in the mind and then provides the user \\ with the answer.
		The reasoning process and answer are enclosed within <think> </think> and \\<answer> </answer> tags, respectively, i.e., <think> reasoning process here </think> \\ <answer> answer here </answer>. \\
		
		Please reason step by step, and put your final answer within \verb|\|boxed\{\}.\\
		
		User: \textbf{\{{question\}}} \\
		Assistant: \\
		\bottomrule
	\end{tabular}
	\label{template}
\end{table}

\subsection{Training Dataset Curation}
To select and curate high-quality data for scaling RL, we include challenging problems from DeepScaleR~\cite{deepscaler}, DAPO-Math-17K~\cite{dapo}, and MATH~\cite{dataset_math} to enhance problem difficulty and diversity in our data mixture:

\noindent{\textbf{DeepScaleR}}\footnote{\url{https://huggingface.co/datasets/agentica-org/DeepScaleR-Preview-Dataset}}, which contains approximately 40K unique mathematics-specific problem-answer pairs collected from AIME (1984-2023), AMC (prior to 2023), Omni-MATH, and Still datasets~\cite{dataset_minerva,OmniMATH,still_dataset}.

\noindent{\textbf{DAPO-Math-17K}}\footnote{\url{https://huggingface.co/datasets/BytedTsinghua-SIA/DAPO-Math-17k}}, which contains approximately 17K problem-answer pairs, each paired with an integer as the answer. DAPO-Math-17K was compiled from the Art of Problem Solving (AoPS\footnote{\url{https://artofproblemsolving.com/}}) website and official competition websites using a combination of web scraping and manual annotation.

\noindent{\textbf{MATH}}\footnote{\url{https://huggingface.co/datasets/EleutherAI/hendrycks_math}} (Level 3-5), which contains approximately 8K problem-answer pairs. Each problem has a step-by-step solution which can be used to teach models to generate explanations.

After obtaining the above datasets, we employ Math-Verify\footnote{\url{https://github.com/huggingface/Math-Verify}} to re-extract answers from the provided textual solutions, selecting only those cases where the extracted answer matches the corresponding answer in the dataset. We discard any samples that are empty, incomplete, or duplicates. Finally, we obtain approximately 59K reasoning problems as the training dataset. It should be noted that in the first stage, we use the 59K data to incentivize the model's reasoning ability. Still, in the second stage, we use the MATH (Level 3-5) data as the training set to optimize the model's reasoning length.
\begin{table}[t!]
	\centering
	\caption{Overall performance on five competition-level reasoning benchmarks. Our models outperform prior state-of-the-art approaches with zero RL paradigm. $^\dagger$ indicate the results from~\cite{sober}.}
	\label{main_results}
	\resizebox{\textwidth}{!}{
		\begin{tabular}{lcccccc}
			\toprule
			\textbf{Model} & \textbf{AIME 2024} & \textbf{MATH-500} & \textbf{AMC 2023}  & \textbf{Minerva} & \textbf{Olympiad} & \textbf{Avg. Score}  \\\midrule
			
			Qwen2.5-1.5B-Base~\cite{qwen2.5}$^\dagger$  & 0.0 & 3.3 & 2.5 & 1.8 & 1.5 & 1.82\\
			Qwen2.5-1.5B-Instruct~\cite{qwen2.5}$^\dagger$  & 1.3 & 57.5 & 26.2 & 19.4 & 20.3& 24.9 \\
			Qwen2.5-Math-1.5B-Base~\cite{qwen2.5_math}$^\dagger$  & 11.3 & 51.7 & 44.0 & 11.3 & 26.0 & 28.9\\
			Qwen2.5-Math-1.5B-Instruct~\cite{qwen2.5_math}$^\dagger$  & 12.0 & 74.7 & 26.7 & 35.0 & 37.9 & 37.3 \\
			DeepSeek-R1-Distill-Qwen-1.5B~\cite{r1}  & 28.8 & 82.8  & 62.9 & 26.5  & 43.3  &48.9 \\
			DeepScaleR-1.5B-Preview~\cite{deepscaler}  & 43.1 & 87.8  & 73.6 & 30.2  & 50.0  & 56.9 \\
			FastCuRL-1.5B-Preview~\cite{fastcurl}  & 43.1 & 88.0 & 74.2 & 31.6 & 50.4 & 57.5\\
			FastCuRL-1.5B-V3~\cite{fastcurl}  & 49.6 & 90.5 & 78.5 & 34.7 & 54.5 & 61.6\\
			
			\midrule\midrule
			Qwen2.5-7B-Base~\cite{qwen2.5}$^\dagger$  & 3.3 & 64.6 & 30.0 & 25.7 & 29.0 & 30.5\\
			Qwen2.5-7B-Instruct~\cite{qwen2.5}$^\dagger$  & 12.3 & 77.1 & 52.8 & 34.9 & 38.7 & 43.2 \\
			Qwen2.5-Math-7B-Base~\cite{qwen2.5_math}$^\dagger$  & 20.7 & 64.3 & 56.2 & 17.3 & 29.0 & 37.5\\
			Qwen2.5-Math-7B-Instruct~\cite{qwen2.5_math}$^\dagger$  & 15.7 & 82.9 & 67.0 & 35.0 & 41.3& 48.4 \\
			Eurus-2-7B-PRIME~\cite{prime}$^\dagger$ & 17.8 & 80.1 & 63.0 & 37.5 & 43.9 & 48.5 \\
			Open-Reasoner-Zero-7B~\cite{orz}$^\dagger$ & 19.7 &  83.9 & 59.5  & 31.6 & 47.6 & 48.5 \\
			SimpleRL-Zero-7B~\cite{zeng2025simplerl}$^\dagger$ & 14.0 & 77.9 & 58.0 & 33.0 & 39.0 &44.4 \\
			SimpleRL-Zero-Math-7B~\cite{zeng2025simplerl}$^\dagger$ & 22.7 & 76.9 & 62.2 & 30.1 & 39.3 &46.2 \\
			Oat-Zero-7B~\cite{drgrpo}$^\dagger$  & 28.0 & 79.4 & 66.2 & 34.4 & 43.8 & 50.4  \\
			\midrule
			\textbf{ConciseR-Zero-7B-Preview} (Stage-1) & {42.8}  & {83.0} & {73.9} & {31.8} & {45.1} & {55.3} \\
			\textbf{ConciseR-Zero-7B} (Stage-2)  & {43.3}   & {83.0} & {76.7} & 31.5 & {46.0} & {56.1}\\
			\bottomrule
	\end{tabular}}
\end{table}
\begin{figure}[t!]
	\begin{center}
		\includegraphics[scale=0.37]{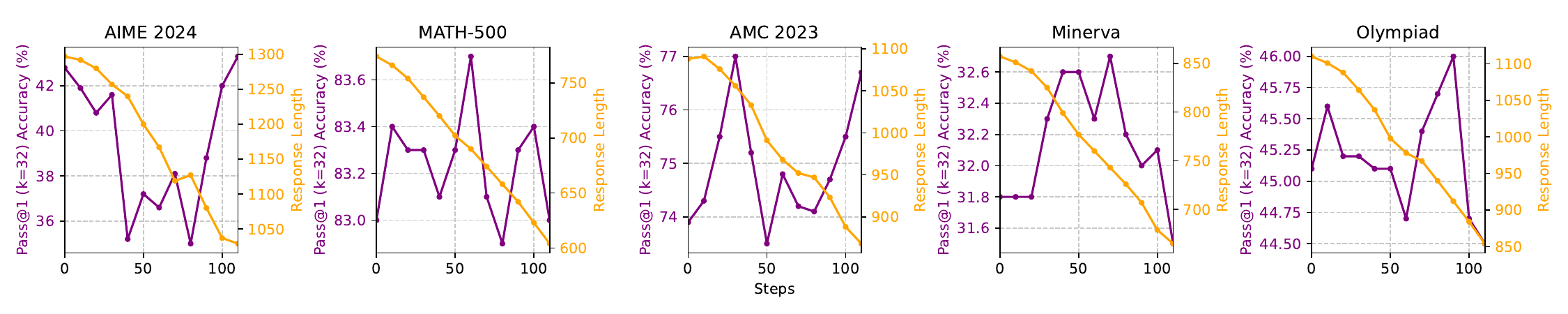}
	\end{center}
	\caption{A detailed evaluation of accuracy and response length throughout the training steps.}\label{l-grpo}
\end{figure}

\section{Experiments}\label{experiments}
\noindent{\textbf{Training Details}}. In this paper, we train our models using the verl\footnote{\url{https://github.com/volcengine/verl}} framework~\cite{verl} and leverage Qwen2.5-Math-7B~\cite{qwen2.5_math} as the base model. During training, we utilize the Adam [39] optimizer with a constant learning rate of $1\times10^{-6}$. We leverage a batch size of 128 with each question generating 32 rollouts, the maximum response length is set to 3,072 tokens, and training is conducted using mini-batches of size 128. As for the Clip-Higher, similar to the prior work~\cite{dapo}, we set the clipping parameter $\varepsilon_l$ to 0.2 and $\varepsilon_h$ to 0.28, which effectively balance the trade-off between exploration and exploitation for RL. Specifically, for GRPO++, we set the entropy coefficient $\alpha$ to 0.001. For L-GRPO, we set the KL penalty coefficient $\beta$ to 0.01 and set the $\lambda$ to 0.000002. 

\noindent{\textbf{Evaluation Benchmarks}}. Similar to the prior work~\cite{drgrpo,fastcurl}, the performance of our models is evaluated on a diverse suite of competition-level benchmarks including AIME 2024\footnote{\url{https://huggingface.co/datasets/AI-MO/aimo-validation-aime}} (comprises 30 challenge problems), AMC 2023\footnote{\url{https://huggingface.co/datasets/AI-MO/aimo-validation-amc}} (contains 40 mathematical problems, covering algebra, geometry, number theory, and combinatorics), Minerva Math~\cite{dataset_minerva}, MATH-500~\cite{dataset_math} (is a challenging benchmark comprising competition-level problems), and OlympaidBench~\cite{dataset_olympiad}.

\noindent{\textbf{Evaluation Setup}}. In this paper, our two-stage RL training framework aims to enhance the reasoning performance while reducing the response length, thereby enabling more concise reasoning. To this end, we adopt the Pass@k evaluation metric, reporting Pass@1 accuracy computed with a non-zero sampling temperature. Therefore, we set the maximum response length to 3,072 tokens. Specifically, we select a temperature of $0.6$ combined with a top-p value of $0.95$ to generate multiple responses (typically 32 samples) for each query. The used training template is shown in Figure~\ref{template}.


\noindent{\textbf{Baselines}}. We conduct comprehensive evaluations against several baselines with zero RL paradigm, including Qwen2.5~\cite{qwen2.5}, Qwen2.5-Math~\cite{qwen2.5_math}, SimpleRL-Zero~\cite{zeng2025simplerl}, Open-Reasoner-Zero-7B~\cite{orz}, Eurus-2-7B-PRIME~\cite{prime}, and Oat-Zero-7B~\cite{drgrpo}. Furthermore, we also present the results of the models after RL based on the DeepSeek-R1-Distill-Qwen-1.5B model~\cite{r1}, including DeepScaleR-1.5B-Preview~\cite{deepscaler}, FastCuRL-1.5B-Preview, and FastCuRL-1.5B-V3~\cite{fastcurl}.

\subsection{Main Results}
The experimental results reported in Table~\ref{main_results} clearly demonstrate that our proposed model, ConciseR, significantly outperforms existing baselines with zero RL paradigm on five widely recognized reasoning benchmarks. Specifically, ConciseR achieves an average accuracy improvement of 55.2\% compared to the base model, Qwen2.5-Math-7B. Meanwhile, our method, GRPO++, also consistently surpasses all baselines, showing superior overall performance averaged across the five benchmarks.

Figure~\ref{l-grpo} illustrates changes in accuracy and response length during the training process of L-GRPO across the five benchmarks. As indicated, the average accuracy on each benchmark remains stable throughout training, without exhibiting any noticeable degradation. Interestingly, the average response length on each benchmark consistently decreases, with reductions of 21\%, 22\%, 20\%, 22\%, and 23\% observed on AIME 2024, MATH-500, AMC 2023, Minerva, and Olympiad benchmarks, respectively. This demonstrates that our training approach successfully maintains model accuracy while generating more concise and efficient responses.

\begin{figure}[h]
	\begin{center}
		\includegraphics[scale=0.27]{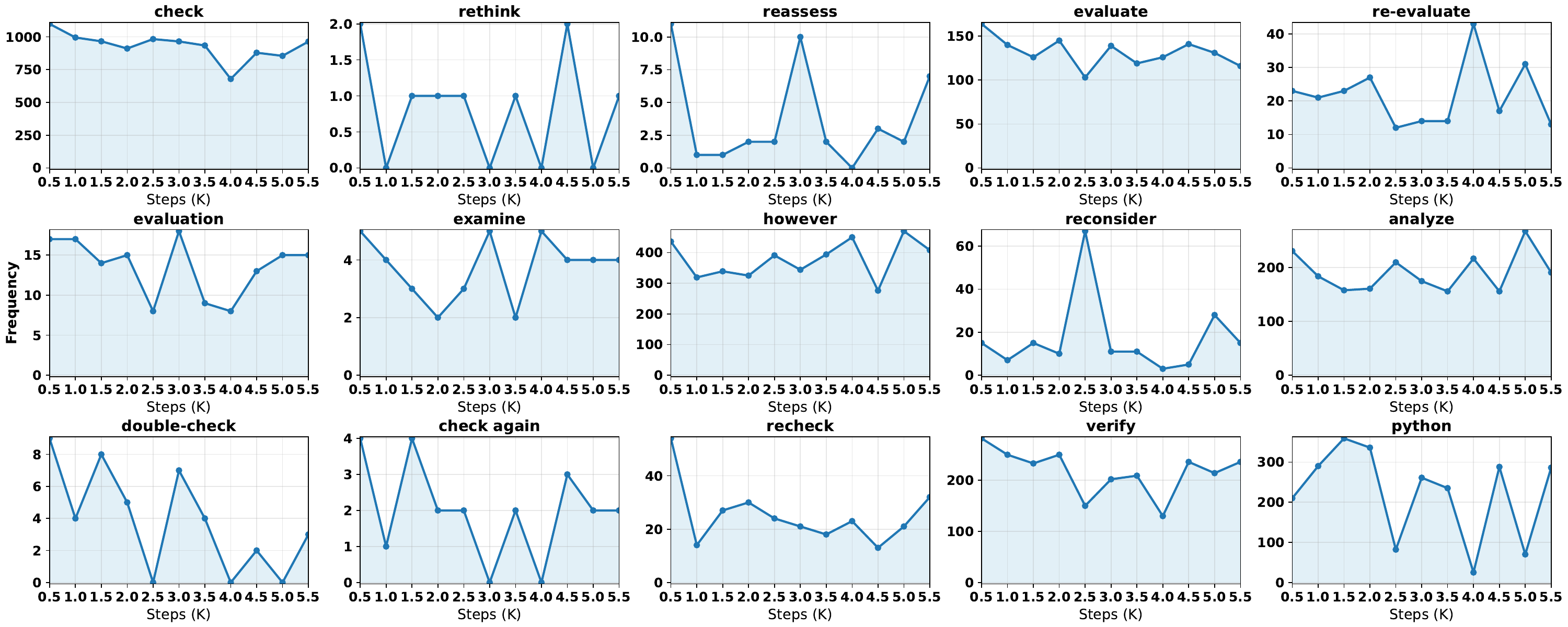}
	\end{center}
	\caption{Count of keyword occurrences out of 14,022 responses (1558 questions $\times$ 11 test times).}\label{stage1}
\end{figure}
\begin{figure}[t!]
	\begin{center}
		\includegraphics[scale=0.27]{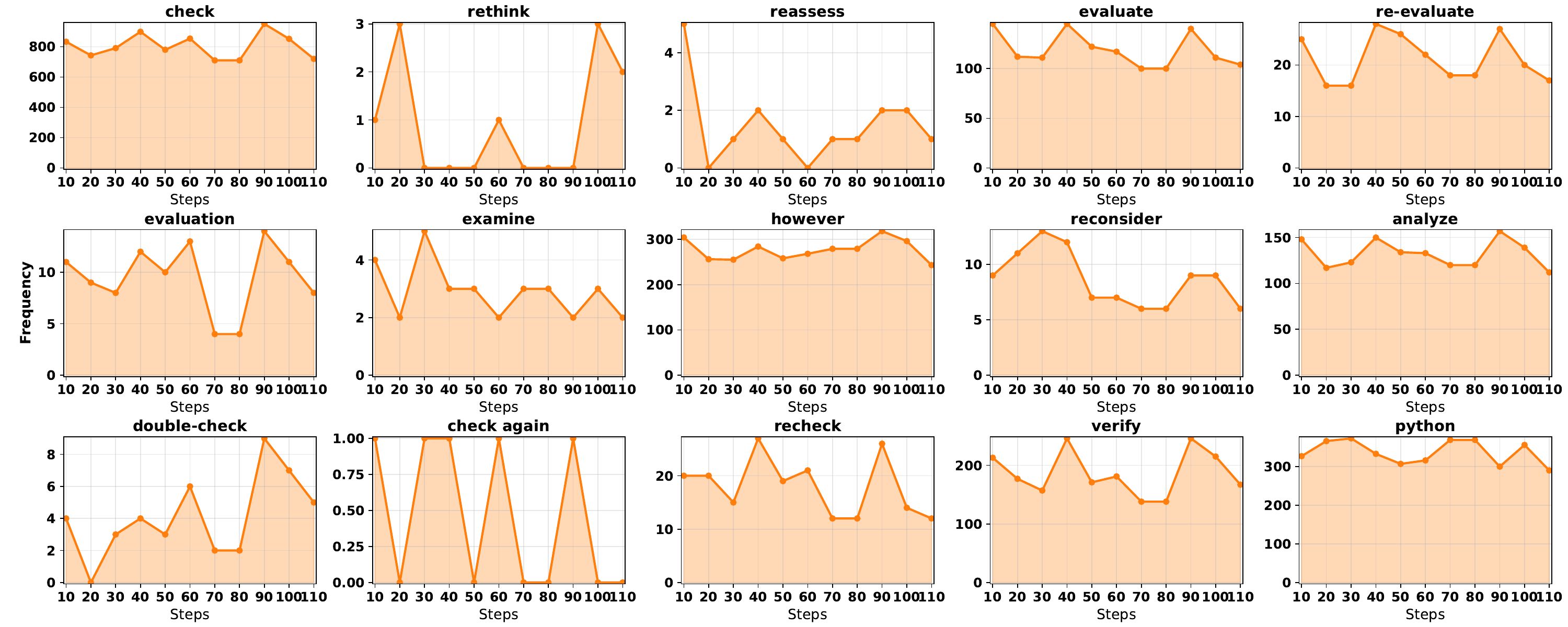}
	\end{center}
	\caption{Count of keyword occurrences out of 15,580 responses (1558 questions $\times$ 11 test times).}\label{stage2}
\end{figure}
\section{Discussions}\label{discussions}
\subsection{Analysis of Changes in Reasoning Patterns}
Inspired by prior work that observes the model's reflective behavior by constructing a keyword pool, we have built a carefully selected keyword pool to observe changes in the thinking patterns of the responses during training. 
In our experiment, the keyword pool is limited to: check,  rethink, reassess, evaluate, re-evaluate, evaluation, examine, however, reconsider,  analyze, double-check, check again, recheck, verify, and wait.
Then, we present the occurrences of various keywords in the responses generated by different training stages and steps in Figure~\ref{stage1} and Figure~\ref{stage2}. Interestingly, when comparing the first and second stages, the frequency with which the model uses code to verify results has significantly increased (as reflected in the frequency of the keyword "python"). The model may have discovered that verifying results by writing code is more efficient. Meanwhile, keywords like "re-check" have decreased relatively, and other keywords have remained unchanged.

\subsection{Case Study}
\begin{figure}[t!]
	\begin{center}
		\includegraphics[scale=0.37]{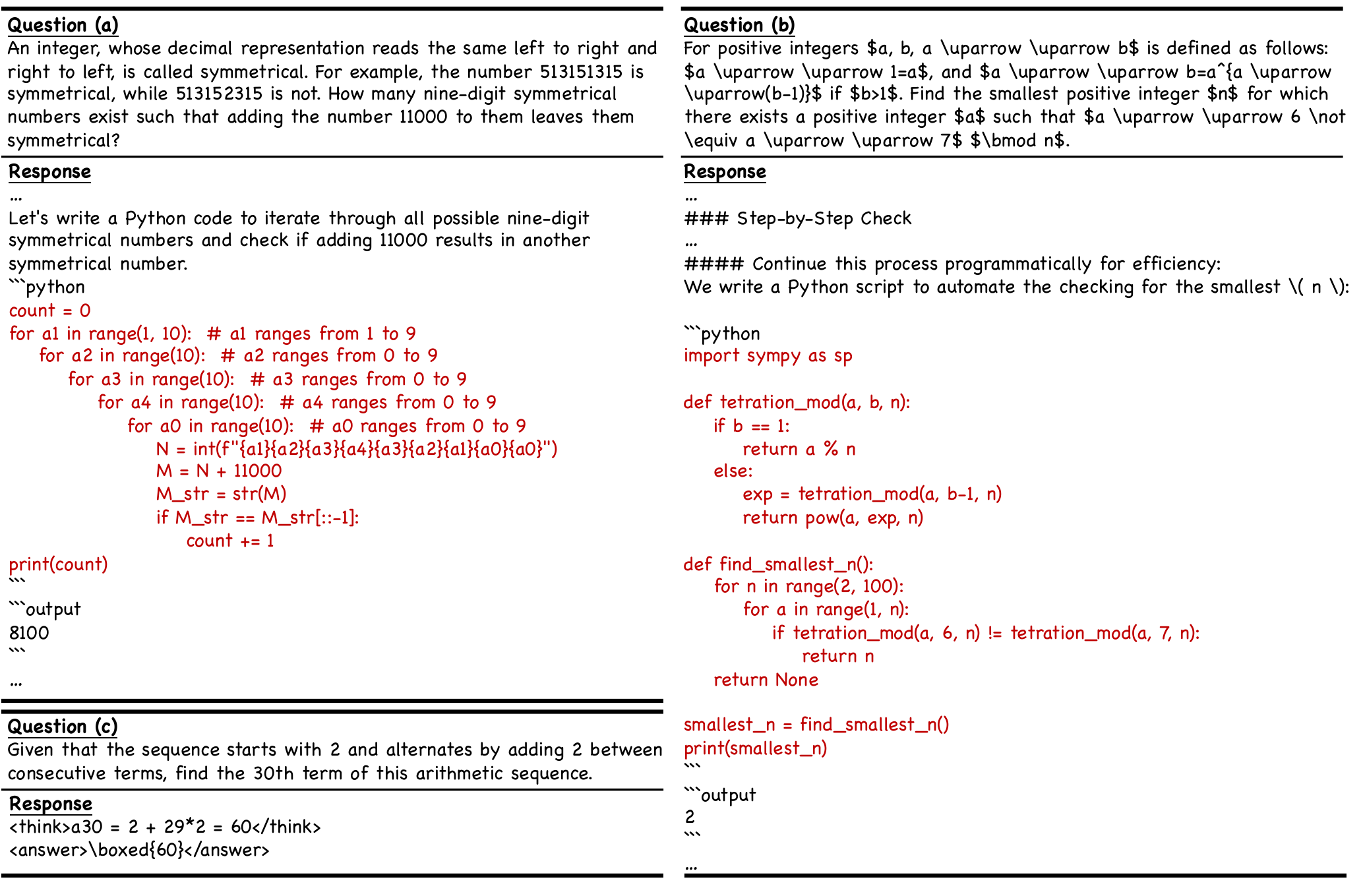}
	\end{center}
	\caption{Illustration of cases.}\label{case}
\end{figure}

An interesting observation is that python code is used for verification during mathematical problem solving, e.g., Questions (a) and (b) in Figure~\ref{case}. Specifically, for Question (a), the model utilizes program code to calculate the answer. For Question (b), the model first presents the solution process through mathematical reasoning and then spontaneously writes program code to verify the correctness of the approach. Such cases illustrate how models employ procedural reasoning to self-correct and engage in subsequent attempts.

\subsection{Failure Experience}
In this section, we discuss our failure experiences in reward shaping. These experiments could also be regarded as an ablation study in L-GRPO. During the initial design of L-GRPO, we consider directly comparing the generation length of samples within a group, assigning higher rewards to samples with relatively shorter CoT reasoning responses. We then combine the length score with the accuracy reward to encourage the trained model to obtain correct answers through shorter CoT reasoning responses, as illustrated by the following two equations,
\begin{equation}
	\hat{r}_i = r_i + \lambda\hat{\mathcal{L}}_i, \quad \hat{\mathcal{L}}_i = 
	\begin{cases}
		\frac{\text{Max}(\{\mathcal{L}_1, \mathcal{L}_2, ..., \mathcal{L}_G\}) - \mathcal{L}_i }{\text{Max}(\{\mathcal{L}_1, \mathcal{L}_2, ..., \mathcal{L}_G\}) - \text{Min}(\{L_1, \mathcal{L}_2, ..., \mathcal{L}_G\})}, & \text{if}\ \sum_{i=1}^{\text{G}}r_i = \text{G} \\
		0, & \text{if}\ \sum_{i=1}^{\text{G}}r_i \neq \text{G}
	\end{cases},
\end{equation}
\begin{equation}
	\hat{r}_i = \frac{\text{G}_{r_i\neq0}-1}{\text{G}_{r_i\neq0}}r_i+\frac{\hat{\mathcal{L}}_i}{\sum_{i=1}^{\text{G}}\hat{\mathcal{L}}_i}, \quad \hat{\mathcal{L}}_i = 
	\begin{cases}
		0, & \text{if}\ \text{is\_equivalent}(o_i, a) \\
		1 - \frac{\mathcal{L}_i}{\sum_{i=1}^{\text{G}}\mathcal{L}_i}, & \text{if}\ \text{not is\_equivalent}(o_i, a)
	\end{cases}.
\end{equation}
However, we find that this direct rewarding easily causes the model to skip the reasoning process and immediately start guessing answers, manifesting as an empty reasoning response within the <think></think> tags while directly outputting the final answer within the <answer></answer> tags. On the contrary, indirectly using the maximum context length to design the reward function can, to some extent, avoid the issues mentioned above.

\section{Related Work}
Recent advances in reinforcement learning have significantly enhanced the reasoning capabilities of large language models. A pivotal development in this domain is OpenAI's o1~\cite{o1}, which leverages large-scale RL training to promote CoT reasoning. This approach has resulted in notable improvements in complex mathematics and coding benchmarks. DeepSeek-R1~\cite{r1} demonstrates that pure RL post-training via GRPO, without the need for supervised warm-up, can directly induce robust reasoning abilities. Remarkably, this kind of method not only achieves performance competitive with o1 but also exhibits emergent behaviors such as self-verification and multi-step planning. This paradigm shift significantly reduces memory and computational overhead compared to earlier GRPO implementations~\cite{orz, zeng2025simplerl, openr1}, all while maintaining competitive performance levels. 
Recent algorithmic variants have focused on enhancing training efficiency \cite{deepscaler,kimik1.5,fastcurl,dapo,drgrpo,fatemi2025concise,zeng2025simplerl,wen2025light}, yet they preserve GRPO's core methodology of parallel CoT sampling across groups.
These advancements collectively contribute to more efficient and robust training methodologies for LLMs, thereby enhancing their reasoning capabilities and performance on complex tasks.

\section{Conclusion}
In this paper, we propose ConciseR, which introduces a simple yet effective two-stage reinforcement learning framework. First, it incentivizes the model's reasoning capabilities via GRPO++, and then it reduces the model's response length to improve the quality of the CoT response implicitly via L-GRPO. \textit{Importantly, we innovatively propose that during training, response length optimization is only triggered when all rollouts for a given training sample are correct. This embodies the "walk before you run" principle.}
Experiments demonstrate that ConciseR consistently achieves the best efficiency-accuracy synergistic improvement, significantly outperforming existing efficient reasoning methods across five benchmarks.

\clearpage
\bibliographystyle{unsrt}
\bibliography{reference}
\end{document}